\documentclass[sigconf,usenames,dvipsnames]{acmart}

\renewcommand\footnotetextcopyrightpermission[1]{} 
\usepackage{booktabs} 
\usepackage{enumitem}
\usepackage{xspace}
\usepackage{xcolor}
\usepackage{multirow}

\newcommand{\hide}[1]{}

\newcommand{\ourmethod}{\textsc{Shield}\xspace}

\newcommand{\gap}{\vspace{4pt}}
\newcommand\Mark[1]{\textsuperscript#1}

\let\nocompression\varnothing

\newcommand{\df}[1]{DeepFool}

\begin{document}
\title[\ourmethod: Fast, Practical Defense \& Vaccination for Deep Learning via JPEG Compression]{\ourmethod: 
Fast, Practical Defense and Vaccination for Deep Learning using JPEG Compression
}





\author{Nilaksh Das\Mark{1}, Madhuri Shanbhogue\Mark{1}, Shang-Tse Chen\Mark{1}, Fred Hohman\Mark{1}, Siwei Li\Mark{1}, Li Chen\Mark{2}, Michael E. Kounavis\Mark{2}, Polo Chau\Mark{1}}
\orcid{0000-0002-5281-5549}
\affiliation{
  \institution{\Mark{1}Georgia Institute of Technology, Atlanta, GA, USA}
}
\email{{nilakshdas, madhuri.shanbhogue, schen351, fredhohman, robertsiweili, polo}@gatech.edu}
\affiliation{
  \institution{\Mark{2}Intel Corporation, Hillsboro, OR, USA}
}
\email{{li.chen, michael.e.kounavis}@intel.com}




 


\renewcommand{\shortauthors}{Das et al.}

\begin{abstract}
The rapidly growing body of research in adversarial machine learning has demonstrated that deep neural networks (DNNs) are highly vulnerable to adversarially generated images.
%
This underscores the urgent need for practical defense techniques
that can be readily deployed to combat attacks in real-time.
%
Observing that many attack strategies aim to perturb image pixels in ways that are visually imperceptible,
we place JPEG compression at the core of our proposed \ourmethod defense framework, utilizing its capability to effectively ``compress away'' such pixel manipulation.
To immunize a DNN model from artifacts introduced by compression, \ourmethod ``vaccinates'' the model by retraining it with compressed images, where different compression levels are applied to generate multiple vaccinated models that are ultimately used together in an ensemble defense.
On top of that, \ourmethod adds an additional layer of protection by employing randomization at test time that compresses different regions of an image using random compression levels, making it harder for an adversary to estimate the transformation performed.
This novel combination of vaccination, ensembling, and randomization makes \ourmethod a fortified multi-pronged defense.
We conducted extensive, large-scale experiments using the ImageNet dataset, and show that our approaches eliminate up to 94\% of black-box attacks and 98\% of gray-box attacks delivered by the recent, strongest attacks, such as \textit{Carlini-Wagner's L2} and \textit{DeepFool}.
Our approaches are fast and work without requiring knowledge about the model.
To enable reproducibility of our results, we have open-sourced our code on GitHub (\url{https://github.com/poloclub/jpeg-defense}).
\end{abstract}

%
%


\maketitle

\begin{figure}[t!]
    \centering
    \includegraphics[width=\linewidth]{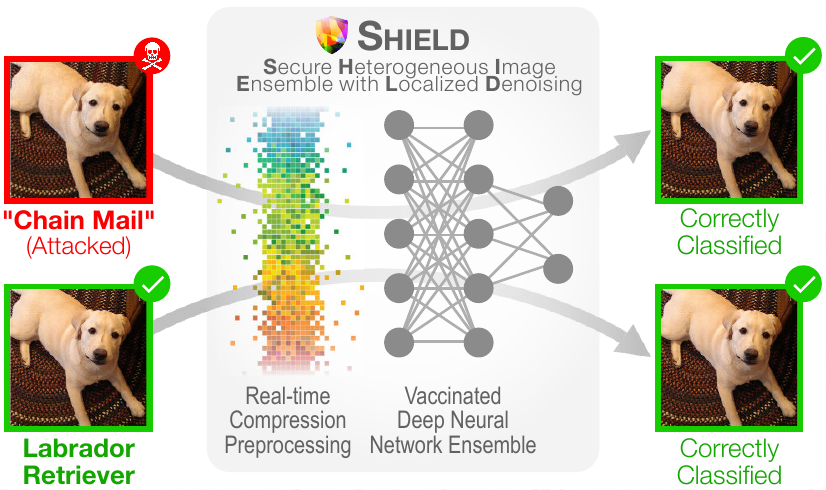}
    \caption{
    \ourmethod Framework Overview. 
    \ourmethod combats adversarial images (in red), by removing perturbation in real time 
    using Stochastic Local Quantization (SLQ) and an ensemble of vaccinated models robust to compression transformation for both adversarial and benign images.
    Our approach eliminates up to 94\% of black-box attacks and 98\% of gray-box attacks delivered by some of the most recent, strongest attacks, such as \textit{Carlini-Wagner's L2} and \textit{DeepFool}.
    }
    \label{fig:crownjewel}
\end{figure}

\section{Introduction}
\label{sec:intro}

Deep neural networks (DNNs), while enjoying tremendous success in recent years, suffer from serious vulnerabilities to adversarial attacks \cite{Szegedy14}.
For example, in computer vision applications, an attacker can add visually imperceptible perturbations to an image and mislead a DNN model into making arbitrary predictions.
When the attacker has complete knowledge of a DNN model, these perturbations can be computed by using the gradient information of the model, which guides the adversary in discovering vulnerable regions 
of the input space that would most drastically affect the model output \cite{goodfellow2014explaining, Papernot16limitation}. 
But even in a black-box scenario, where the attacker does not know the exact network architecture, one can use a substitute model to craft adversarial perturbations that are transferable to the target model \cite{Papernot17blackbox}.
To make this even more troubling, it is possible to print out physical 2D or 3D objects to fool recognition systems in realistic settings \cite{sharif2016accessorize, athalye2017synthesizing}.

The threat of adversarial attack casts a shadow over deploying DNNs
in security and safety-critical applications like self-driving cars.
To better understand and fix the vulnerabilities, there is a growing body of research on defending against various attacks and making DNN models more robust \cite{papernot2016distillation, bhagoji2017dimensionality, metzen2017detecting}. 
However, the progress of defense research has been lagging behind the attack side so far. 
Moreover, research on defense rarely focuses on practicality and scalability, both essential for real-world deployment. 
For example, total variation denoising and image quilting are image preprocessing techniques that have potential in mitigating adversarial perturbations to some extent \cite{guo2018countering}, but they incur significant computational overhead, calling into question how feasibly they can be used in practical applications, which often require fast, real-time defense \cite{evtimov2017robust, eykholt2017note}.

\subsection{Our Contributions and Impact}

\noindent \textbf{1. Compression as Fast, Practical, Effective Defense.}
We contribute the major idea that \textit{compression} --- a central concept that underpins numerous successful data mining techniques --- can offer powerful, scalable, and practical protection for deep learning models against adversarial image perturbations in real-time.
Motivated by our observation that many attack strategies aim to perturb images in ways that are visually imperceptible to the naked eye, 
we show that systematic adaptation of the widely available JPEG compression technique can effectively compress away such  pixel ``noise'', especially since JPEG is particularly designed to reducing image details that are imperceptible to humans.
(Section \ref{ssec:preprocess})

\medskip
\noindent \textbf{2. \ourmethod: Multifaceted Defense Framework.} 
Building on our foundational compression idea, we contribute the novel \ourmethod defense framework that combines \textit{randomization}, \textit{vaccination} and \textit{ensembling} into a fortified multi-pronged protection:

\begin{enumerate}[itemsep=1pt, topsep=3pt, partopsep=0pt, leftmargin=15pt]

\item We exploit JPEG's flexibility in supporting varying compression levels to develop strong ensemble models that span a spectrum of compression levels; 

\item We show that a model can be ``vaccinated'' by training on compressed images, increasing its robustness towards compression transformation for both adversarial and benign images;

\item 
\ourmethod employs stochastic quantization that compresses different regions of an image using randomly sampled compression levels, making it harder for the adversary to estimate the transformation performed.

\end{enumerate}

\noindent \ourmethod does not require any change in the model architecture, and can recover significant amount of model accuracy lost to adversarial instances, with little effect on the accuracy for benign instances.
\ourmethod stands for \textbf{S}ecure \textbf{H}eterogeneous \textbf{I}mage \textbf{E}nsemble with \textbf{L}ocal \textbf{D}enoising.
To the best of our knowledge, our multi-faceted defense approach has yet been challenged.
(Sections \ref{ssec:vaccination} \& \ref{ssec:shield})

\medskip
\noindent \textbf{3. Extensive Evaluation Against Major Attacks.}
We perform extensive experiments using the full ImageNet benchmark dataset with 50K images, demonstrating that our approach is fast, effective and scalable. 
Our approaches eliminate up to 94\% of black-box attacks and 98\% of gray-box attacks delivered by some of the most recent, strongest attacks, such as \textit{Carlini-Wagner's L2} \cite{carlini2017towards} and \textit{DeepFool} \cite{Moosavi16}.
(Section \ref{sec:eval})

\medskip
\noindent \textbf{4. Impact to Intel and Beyond.}
This work is making multiple positive impacts on Intel's research and product development plans.
Introduced with the Sandy Bridge CPU microarchitecture, 
Intel's Quick Sync Video (QSV) technology dedicates a hardware core for high-speed video processing,
performs JPEG compression up to 24x faster than TensorFlow implementations,
paving the way for real-time defense in safety-critical applications, such as autonomous vehicles.
This research has sparked insightful discussion among research and development teams at Intel, on the priority of \textit{secure deep learning} that necessitates tight integration of practical defense strategies, software platforms and hardware accelerators. 
We believe our work will accelerate the industry's emphasis on this important topic.
To ensure reproducibility of our results, we have open-sourced our code on GitHub (\url{https://github.com/poloclub/jpeg-defense}).
(Section \ref{sec:impact})

\section{Background: Adversarial Attacks}
\label{sec:background}


Our work focuses on defending against adversarial attacks on deep learning models.
This section provides  background information for readers new to the adversarial attack literature. 


Given a trained classifier $C$ and an instance $x \in \mathcal{X}$, the objective of an adversarial \textit{untargeted} attack is to compute a perturbed instance $x'$ such that $C(x') \ne C(x)$ and $d(x, x') \leq \rho$ for some distance function $d(\cdot, \cdot)$ and $\rho\ge 0$.
Popular choices of $d(\cdot, \cdot)$ are Euclidean distance $d(x, x') = \|x - x'\|_2$, and Chebychev distance $d(x, x') = \|x - x'\|_\infty$.
A \textit{targeted} attack is similar, but is required to induce a 
classification for a specific target class $t$, i.e., $C(x') = t$. In both cases, depending on whether the attacker has full knowledge of $C$ or not, the attack can be further categorized into \textit{white-box} attack and \textit{black-box} attack. 
The latter is obviously harder for the attacker since less information is known about the model, but has been shown to be possible in practice by relying on the property of transferability from a substitute model to the target model when both of them are DNNs trained using gradient backpropagation ~\cite{Szegedy14, Papernot17blackbox}.



The seminal work by Szegedy et al.~\cite{Szegedy14} proposed the first effective adversarial attack on DNN image classifiers by solving a box-constrained L-BFGS optimization problem and showed that the computed perturbations to the images were indistinguishable to the human eye --- a rather troublesome property for people trying to identify adversarial images. 
This discovery has gained tremendous interest, and many new attack algorithms have been invented~\cite{goodfellow2014explaining, Moosavi16, Moosavi17, Papernot16limitation} and applied to other domains such as malware detection~\cite{grosse2016malware, hu2017generating}, sentiment analysis~\cite{PapernotMSH16}, and reinforcement learning~\cite{lin2017tactics, huang2017adversarial}.
Below, we describe the major, well-studied attacks in the literature, against which we will evaluate our approach.

\medskip
\noindent \textbf{Carlini-Wagner's $L_2$} (\emph{CW-L2})~\cite{carlini2017towards} is an optimization-based attack that adds a relaxation term to the perturbation minimization problem based on a differentiable surrogate of the model. They pose the optimization as minimizing:
\begin{align}
\begin{split}
    \|x - x'\|_2 + \lambda \max \big( &- \kappa, Z(x')_k - \max\{Z(x')_{k'}: k' \ne k\} \big)
\end{split}
\end{align}
where $\kappa$ controls the confidence with which an image is misclassified by the DNN,
and $Z(\cdot)$ is the output from the logit layer (last layer before the softmax function is applied for prediction) of $C$.

\medskip
\noindent \textbf{DeepFool} (\emph{DF})~\cite{Moosavi16}  constructs an adversarial instance under an $L_2$ constraint by assuming the decision boundary to be hyperplanar. The authors leverage this simplification to compute a minimal adversarial perturbation that results in a sample that is close to the original instance but orthogonally cuts across the nearest decision boundary. In this respect, \emph{DF} is an untargeted attack. Since the underlying assumption about the decision boundary being completely linear in higher dimensions is an oversimplification of the actual case, \emph{DF} keeps reiterating until a true adversarial instance is found.
The resulting perturbations are harder for humans to detect compared to perturbations introduced by other attacks.


\medskip
\noindent \textbf{Iterative Fast Gradient Sign Method} (\emph{I-FGSM})~\cite{kurakin2016adversarial} is the iterative version of the \textbf{Fast Gradient Sign Method} (\emph{FGSM})~\cite{goodfellow2014explaining}, which is a fast algorithm that computes perturbations subject to an $L_{\infty}$ constraint. \emph{FGSM} simply takes the sign of the gradient of loss function $J$ w.r.t. the input $x$,
\begin{align}
    x' = x + \epsilon \cdot sign(\nabla J_x (\theta, x, y))
\end{align}
where $\theta$ is the set of parameters of the model and $y$ is the true label of the instance. 
The parameter $\epsilon$ controls the magnitude of per-pixel perturbation. 
\emph{I-FGSM} iteratively applies FGSM in each iteration $i$ after clipping the values appropriately at each step:
\begin{align}
    x^{(i)} = x^{(i-1)} + \epsilon \cdot sign(\nabla J_{x^{(i-1)}} (\theta, x^{(i-1)}, y))
\end{align}

\section{Proposed Method: Compression as Defense}
\label{sec:method}



In this section, we present our compression-based approach for combating adversarial attacks. 
In Section \ref{ssec:preprocess}, we begin by describing the technical reasons why compression can remove perturbation.
As compression would modify the distribution of the input space by introducing some artifacts, in Section \ref{ssec:vaccination}, we propose to ``vaccinate'' the model by training it with compressed images, which increases its robustness towards compression transformation for both adversarial and benign images.
Finally, in Section \ref{ssec:shield}, we present our multifaceted \ourmethod defense framework that combines random quantization, vaccination and ensembling into a fortified multi-pronged defense,
which, to the best of our knowledge, has yet been challenged.


\subsection{Preprocessing Images using Compression} 
\label{ssec:preprocess}

Our main idea on rectifying the prediction of a trained model $C$, with respect to a perturbed input $x'$, is to apply a preprocessing operation $g(\cdot)$ that brings back $x'$ closer to the original benign instance $x$, 
which implicitly aims to make $C(g(x')) = C(x)$. 
Constructing such a $g(\cdot)$ is application dependent.
For the image classification problem, we show that JPEG compression is a powerful preprocessing defense technique. JPEG compression mainly consists of the following steps:
\begin{enumerate}
    \item Convert the given image from \emph{RGB} to $YC_bC_r$ (chrominance + luminance) color space. 
    
    \item Perform spatial subsampling of the chrominance channels, since the human eye is less susceptible to these changes and relies more on the luminance information.
    
    \item Transform $8 \times 8$ blocks of the $YC_bC_r$ channels to a frequency domain representation using Discrete Cosine Transform (DCT).
    
    \item Perform quantization of the blocks in the frequency domain representation according to a quantization table which corresponds to a user-defined quality factor for the image.
\end{enumerate}

The last step is where the JPEG algorithm achieves the majority of compression at the expense of image quality. 
This step suppresses higher frequencies more since these coefficients contribute less to the human perception of the image.
As adversarial attacks do not optimize for maintaining the spectral signature of the image, they tend to introduce more high frequency components which can be removed at this step.
This step also renders the preprocessing stage non-differentiable, which makes it non-trivial for an adversary to optimize against, allowing only estimations to be made of the transformation \cite{shin2017jpeg}.
We show in our evaluation (Section \ref{sec:whitebox}) that JPEG compression effectively removes adversarial perturbation across a wide range of compression levels.

\begin{figure}[bt]
    \centering
    \includegraphics[width=\linewidth]{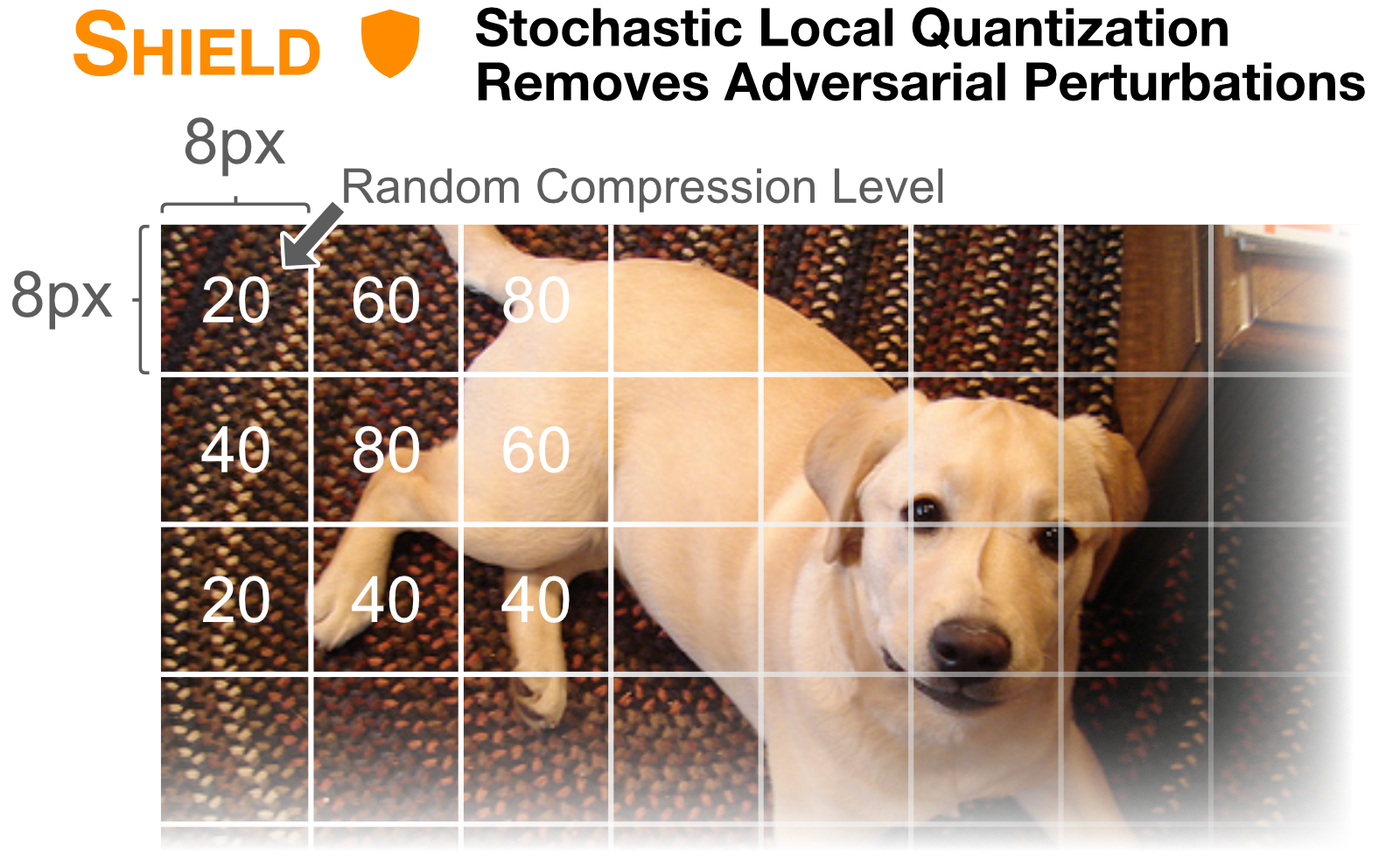}
    \caption{\ourmethod uses Stochastic Local Quantization (SLQ) to remove adversarial perturbations from input images.
    \ourmethod divides images into $8 \times 8$ blocks and applies a randomly selected JPEG compression quality (20, 40, 60 or 80) to each block to remove adversarial attacks.
    Note this figure is an illustration; our images are of actual size $299 \times 299$.
    }
    \label{fig:block}
\end{figure}


\subsection{Vaccinating Models with Compressed Images}
\label{ssec:vaccination}

%
As DNNs are typically trained on high quality images (with little or compression), they are often invariant to the artifacts introduced by the preprocessing of JPEG at high-quality settings.
This is especially useful in an adversarial setting as our pilot study has shown that applying even mild compression removes the perturbations introduced by some attacks \cite{das2017keeping}.
However, applying too much compression could reduce the model accuracy on benign images.


We propose to ``vaccinate'' the model by training it with compressed images, especially those at lower JPEG qualities, which increases the model's robustness towards compression transformation for both adversarial and benign images.
With vaccination, we can apply more aggressive compression to remove more adversarial perturbation.
In our evaluation (Section \ref{sec:graybox}), we show the significant advantage that our vaccination strategy provides, which offers a lift of more than 
7 \textit{absolute} percentage points in model accuracy for high-perturbation attacks.


%

%


\subsection{\ourmethod: Multifaceted Defense Framework}
\label{ssec:shield}

To leverage the effectiveness of JPEG compression as a preprocessing technique along with the benefit of vaccinating with JPEG images, 
we propose a \textit{stochastic variant} of the JPEG algorithm that introduces randomization to the quantization step, making it harder for the adversaries to estimate the preprocessing transformation.

Figure~\ref{fig:block} illustrates our proposed strategy, where we vary the quantization table for each $8 \times 8$ block in the frequency domain to correspond to a random quality factor from a provided set of qualities, such that the compression level does not remain uniform across the image.
This is equivalent to breaking up the image into  disjoint $8 \times 8$ blocks, compressing each block with a random quality factor, and putting the blocks together to re-create the final image.
We call this method \emph{Stochastic Local Quantization} (SLQ).
As the adversary is free to craft images with varying amounts of perturbation,
our defense should offer protection across a wide spectrum.
Thus, we selected the set of qualities $\{20, 40, 60, 80\}$ as our randomization candidates, uniformly spanning the range of JPEG qualities from 1 (most compressed) to 100 (least compressed).
%

Comparing our stochastic approach to taking a simple average over JPEG compressed images, 
our method  allows for maintaining the original semantics of the image in the blocks compressed to higher qualities, while performing more localized denoising in the blocks compressed to lower qualities.
%
In the case of simple average,
all perturbations may not be removed at higher qualities and they might simply dominate the other components participating in the average, still posing to be adversarial. 
Introducing localized stochasticity reduces this expectation.

In our evaluation (Section \ref{sec:graybox}), we will show that by using the spectrum of JPEG compression levels with our stochastic approach,
our model can simultaneously attain a high accuracy on benign images, while being more robust to adversarial perturbations --- a strong benefit that using a single JPEG quality cannot provide.
%
Our method is further fortified by using an ensemble of vaccinated models individually trained on the set of qualities picked for randomization.
We show in Section \ref{sec:graybox} how our method can achieve high model accuracies, comparable to those of much larger ensembles, but is significantly faster.
%

\section{Evaluation}
\label{sec:eval}

In this section, we show that our approach is scalable, effective and practical in removing adversarial image perturbations.
For our experiments, we consider the following scenarios:

\begin{itemize}
    
    \item The adversary has access to the full model, including its architecture and parameters. (Section \ref{sec:whitebox})
    
    \item The adversary has access to the model architecture, but not the exact parameters. (Section \ref{sec:graybox})
    
    \item The adversary does not have access to the model architecture. (Section \ref{sec:blackbox})
\end{itemize}




\begin{figure}[t!b]
    \centering
    \includegraphics[width=\linewidth]{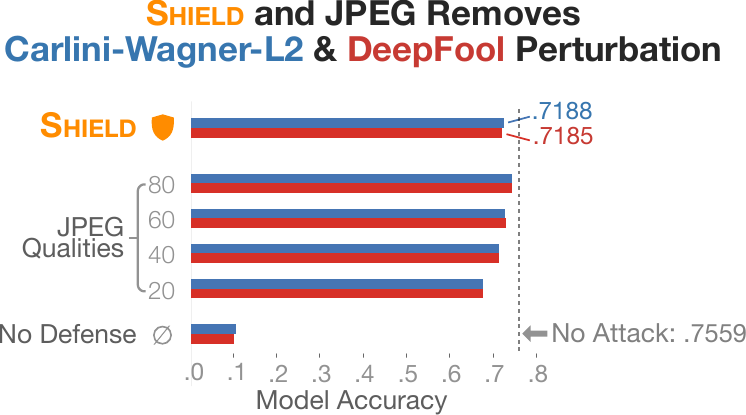}
    \caption{
    Carlini-Wagner-L2 (CW-L2) and DeepFool, two recent strong attacks, introduce perturbations that lowers model accuracy to around 10\% ($\nocompression$). 
    JPEG compression removes up to 98\% of such manipulation (DeepFool), 
    while \ourmethod achieves similar performance, removing up to 94\% of manipulation (DeepFool).
    }
    \label{fig:cwl2-df}
\end{figure}

\subsection{Experiment Setup}

We performed experiments on the full validation set of the \emph{ImageNet} benchmark image classification dataset \cite{krizhevsky2012imagenet}, which consists of 1,000 classes, totaling 50,000 images. 
We show the performance of each defense on the \emph{ResNet-v2 50} model obtained from the \emph{TF-Slim} module in \emph{TensorFlow}. 
We construct the attacks using the popular \emph{CleverHans} package\footnote{\url{https://github.com/tensorflow/cleverhans}}, which contains implementations from the authors of the attacks.

\begin{itemize}
    
    \item For \textit{Carlini-Wagner-L2} (CW-L2), we set its parameter $\kappa = 0$, a common value used in studies \cite{guo2018countering}, as larger values (higher confidence) incur prohibitively high computation cost.

    \item \textit{DeepFool} (DF) is a non-parametric attack that optimizes the amount of perturbation required to misclassify an image. 
    
    \item For \textit{FGSM} and \textit{I-FGSM}, we vary $\epsilon$ from 0 to 8 in steps of 2.  
    
\end{itemize}

\noindent We compare JPEG compression and \ourmethod with two popular denoising techniques that have potential in defending against adversarial attacks~\cite{xu2018feature, guo2018countering}. 
Median filter (MF) collapses a small window of pixels into a single value, and may drop some of the adversarial pixels in the process. 
Total variation denoising (TVD) aims to reduce the total variation in an image, and may undo the artificial noise injected by the attacks.
We vary the parameters of each method to evaluate how their values affect defense performance.


\begin{itemize}

    \item For JPEG compression, we vary the compression level from quality 100 (least compressed) to 20 (greatly compressed), in decrements of 10.
    
    \item For \textit{median filter} (MF), we use window sizes of 3 (smallest possible) and 5. We tested larger window sizes (e.g., 7), which led to extremely poor model accuracies, thus were ruled out as parameter candidates.
    
    \item For \textit{total variation denoising} (TVD), we vary its weight parameter from 10 through 40, in increments of 10. Reducing the weight of TVD further (e.g., 0.3) produces blurry images that lead to poor model accuracy. 
    
\end{itemize}

\begin{figure}[t]
    \centering
    \includegraphics[width=\linewidth]{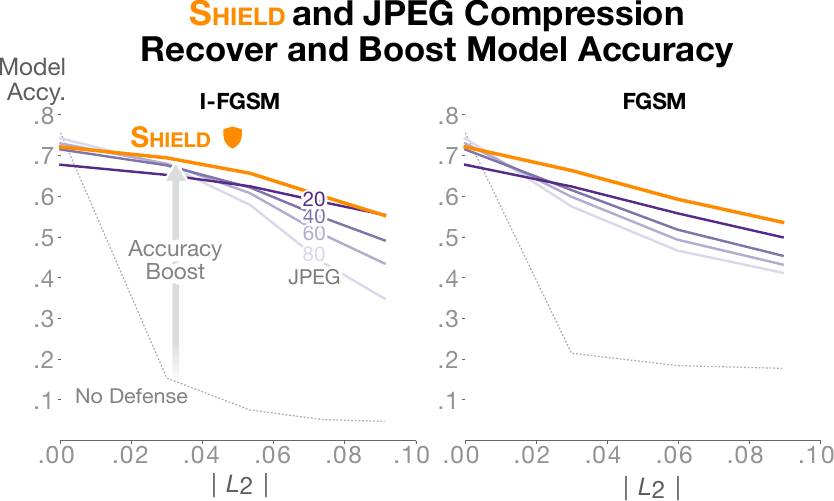}
    \caption{\ourmethod greatly recovers the accuracies for models attacked by I-FGSM (left) or FGSM (right).
    Both charts show a model without defense (gray dotted curve) falls beneath 19\% accuracy.
    However, applying varying JPEG compression qualities (purple curves) helps recover accuracy significantly, and \ourmethod (orange curve) recovers even more accuracy, more than any single JPEG-defended model.}
    \label{fig:ifgsm-fgsm}
\end{figure}


\subsection{Defending Gray-Box Attacks with Image Preprocessing} \label{sec:whitebox}

In this section, we investigate the setting where an adversary gains access to all parameters and weights of a model that is trained on benign images, but is unaware of the defense strategy.
This constitutes a \textit{gray-box} attack on the overall classification pipeline. 

We show the results of applying JPEG compression at various qualities on images attacked with Carlini-Wagner-L2 (CW-L2) and DeepFool (DF) in Figure \ref{fig:cwl2-df}, and on images attacked with I-FGSM and FGSM in Figure \ref{fig:ifgsm-fgsm}.

\medskip
\noindent \textbf{Combating Carlini-Wagner-L2 (CW-L2) \& DeepFool (DF).}
Although CW-L2 and DF, both considered strong attacks, are highly effective at lowering model accuracies, Figure \ref{fig:cwl2-df} shows that even applying mild JPEG compression (i.e., using higher JPEG qualities) can recover much of the lost accuracy.
Since both methods optimize for a lower perturbation to fool the model, the noise introduced by these attacks is imperceptible to the human eye and lies in the high frequency spectrum, which is destroyed in the quantization step of the JPEG algorithm.
\ourmethod performs well, and comparably, for both attacks.
We do not arbitrarily scale the perturbation magnitude of either attack as in \cite{guo2018countering}, as doing so would violate the attacks' optimization criteria.

\medskip
\noindent \textbf{Combating I-FSGM \& FGSM.}
As shown in Figure~\ref{fig:ifgsm-fgsm}, JPEG compression also achieves success in countering I-FGSM and FGSM attacks, which introduce higher magnitudes of perturbation.

As the amount of perturbation increases, the accuracies of models without any protection (gray dotted curves in Figure~\ref{fig:ifgsm-fgsm}) rapidly falls beneath 19\%.
JPEG recovers significant portions of the lost accuracies (purple curves); its effectiveness also gradually and expectantly declines as perturbation becomes severe.
Applying more compression generally recovers more accuracy (e.g., dark purple curve, for JPEG quality 20), 
but at the cost of losing some accuracy for benign images.
\ourmethod (orange curve) offers a desirable trade-off, achieving good performance under severe perturbation while retaining accuracies comparable to the original models. 
Applying less compression (light purple curves) performs well with benign images but is not as effective when perturbation increases.

\begin{table*}[bt]
    \centering
    \begin{tabular}{lrrrrr}
        \toprule
         & No Attack & \textbf{CW-L2} ($\kappa = 0$) & \textbf{DF} & \textbf{I-FGSM} ($\epsilon=4$) & \textbf{FGSM} ($\epsilon=4$)   \\
        Defense & $|L_2| = 0$ & $|L_2| = .0025$ & $|L_2| = .0020$ & $|L_2| = .0533$ & $|L_2| = .0597$  \\
        \midrule
        \emph{No Defense} & \emph{75.59} & \emph{10.29} & \emph{9.78} & \emph{7.49} & \emph{18.40} \\
        \midrule
        \ourmethod [20, 40, 60, 80]  & 72.11 & 71.85 & 71.88 & \textbf{65.63} & \textbf{59.29}\\
        \midrule
        JPEG [quality=100] & \textbf{74.95} & 74.37 & \textbf{74.41} & 52.52 & 44.00 \\
        JPEG [quality=90] & 74.83 & \textbf{74.43} & 74.36 & 55.18 & 45.12 \\
        JPEG [quality=80] & 74.23 & 73.92 & 73.88 & 57.86 & 46.66\\
        JPEG [quality=70] & 73.61 & 73.11 & 73.17 & 59.53 & 47.96 \\
        JPEG [quality=60] & 72.97 & 72.46 & 72.52 & 60.74 & 49.33 \\
        JPEG [quality=50] & 72.32 & 71.86 & 71.91 & 61.47 & 50.53 \\
        JPEG [quality=40] & 71.48 & 71.03 & 71.05 & 62.14 & 51.81 \\
        JPEG [quality=30] & 70.08 & 69.63 & 69.67 & 62.52 & 53.51 \\
        JPEG [quality=20] & 67.72 & 67.32 & 67.34 & 62.43 & 55.81\\
        \midrule
        MF [window=3] & 71.05 & 70.44 & 70.42 & 60.09 & 51.06 \\
        MF [window=5] & 58.48 & 58.19 & 58.06 & 53.59 & 49.71 \\
        \midrule
        TVD [weight=10] & 69.14 & 68.69 & 68.74 & 62.40 & 53.56\\
        TVD [weight=20] & 71.87 & 71.44 & 71.45 & 61.90 & 50.26 \\
        TVD [weight=30] & 72.82 & 72.34 & 72.37 & 60.70 & 48.18 \\
        TVD [weight=40] & 73.31 & 72.90 & 72.91 & 59.60 & 47.07 \\
        \bottomrule
    \end{tabular}
    \caption{
    Summary of model accuracies (in \%) for all defenses: \ourmethod, JPEG, median filter (MF), and total variation denoising (TVD); vs all attacks: Carlini-Wagner L2 (CW-L2), DeepFool (DF), I-FGSM and FGSM. 
    While all techniques are able to recover accuracies from CW-L2 and DF, both strongly optimized attacks with lower perturbation strength, the best performing settings are from JPEG (in bold font).
    \ourmethod benefits from the combination of Stochastic Local Quantization, vaccination and ensembling, outperforming all other techniques when facing high perturbation delivered by I-FGSM and FGSM.
    }
    \label{tab:all-whitebox-accuracies}
\end{table*}

\begin{figure}[b!]
    \centering
    
    \includegraphics[width=\linewidth]{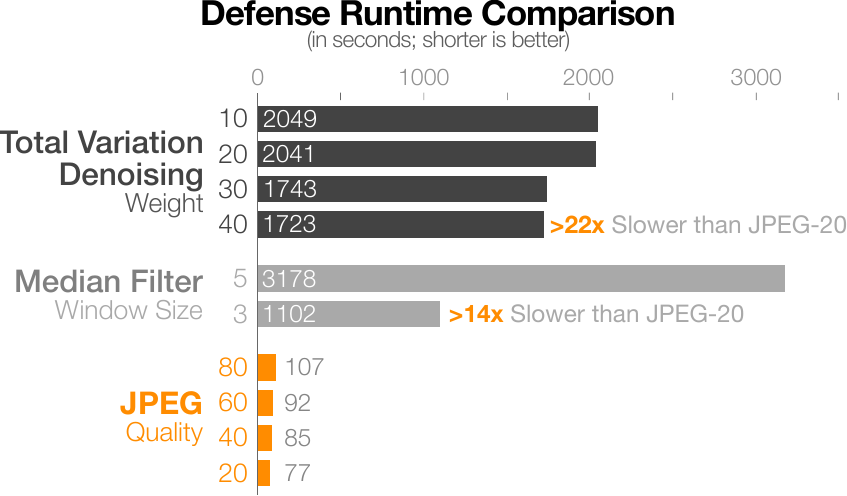}
    \caption{Runtime comparison for three defenses: 
    (1) total variation denoising (TVD), 
    (2) median filter (MF), and 
    (3) JPEG compression, 
    timed using the full 50k ImageNet validation images, averaged over 3 runs. 
    JPEG is at least 22x faster than TVD, and 14x faster than MF.
    (Window size of 3 is the smallest possible for median filter.)}
    \label{fig:runtimes}
\end{figure}

\medskip
\noindent \textbf{Effectiveness and Runtime Comparison against Median Filter (MF) and Total Variation Denoising (TVD).}
We compare JPEG compression and \ourmethod with MF and TVD, two popular denoising techniques, because they too have potential in defending against adversarial attacks~\cite{xu2018feature, guo2018countering}. 
Like JPEG, both MF and TVD are parameterized.
Table~\ref{tab:all-whitebox-accuracies} summarizes the performance of all the image preprocessing techniques under consideration. 
While all techniques are able to recover accuracies from CW-L2 and DF, both strongly optimized attacks with lower perturbation strength,  
the best performing settings are from JPEG (bold font in Table~\ref{tab:all-whitebox-accuracies}).
When faced with large amount of perturbation generated by the I-FGSM and FSGM attacks,
\ourmethod benefits from the combination of Stochastic Local Quantization, vaccination, and ensembling, outperforming all other techniques.


As developing practical defense is our primary goal, effectiveness, while important, is only one part of our desirable solution.
Another critical requirement is that our solution be fast and scalable.
Thus, we also compare the runtimes of the image processing techniques. 
Our comparison focuses on the most computationally intensive parts of each technique, ignoring irrelevant overheads (e.g., disk I/O) common to all techniques.
All runtimes are averaged over 3 runs, using the full 50k ImageNet validation images, on a dedicated desktop computer equipped with an Intel i7-4770K quad-core CPU clocked at 3.50GHz, 4x8GB RAM, 1TB SSD of Samsung 840 EVO-Series and 2x3TB WD 7200RPM hard disk, running Ubuntu 14.04.5 LTS and Python 2.7. 
We used the fastest, most popular Python implementations of the image processing techniques.
We used JPEG and MF from Pillow 5.0, and TVD from scikit-image.

As shown in Figure~\ref{fig:runtimes}, JPEG is the fastest, spending no more than 107 seconds to compress 50k images (at JPEG quality 80).
It is at least 22x faster than TVD, and 14x faster than median filter.
We tested the speed of the TensorFlow implementation of \ourmethod, which also compresses all images at high speed, taking only 150s.

\subsection{Black-Box Attack with Vaccination and Ensembling}
\label{sec:graybox}

We now turn our attention to the setting where an adversary has knowledge of the model being used but does not have access to the model parameters or weights. More concretely, we vaccinate the ResNet-v2 50 model by retraining on the ImageNet training set and preprocessing the images with JPEG compression while training.
This setup constitutes a \textit{black-box} attack, as the attacker only has access to the original model but not the vaccinated model  being used. 

\begin{figure*}[tb]
    \centering
    \includegraphics[width=0.8\textwidth]{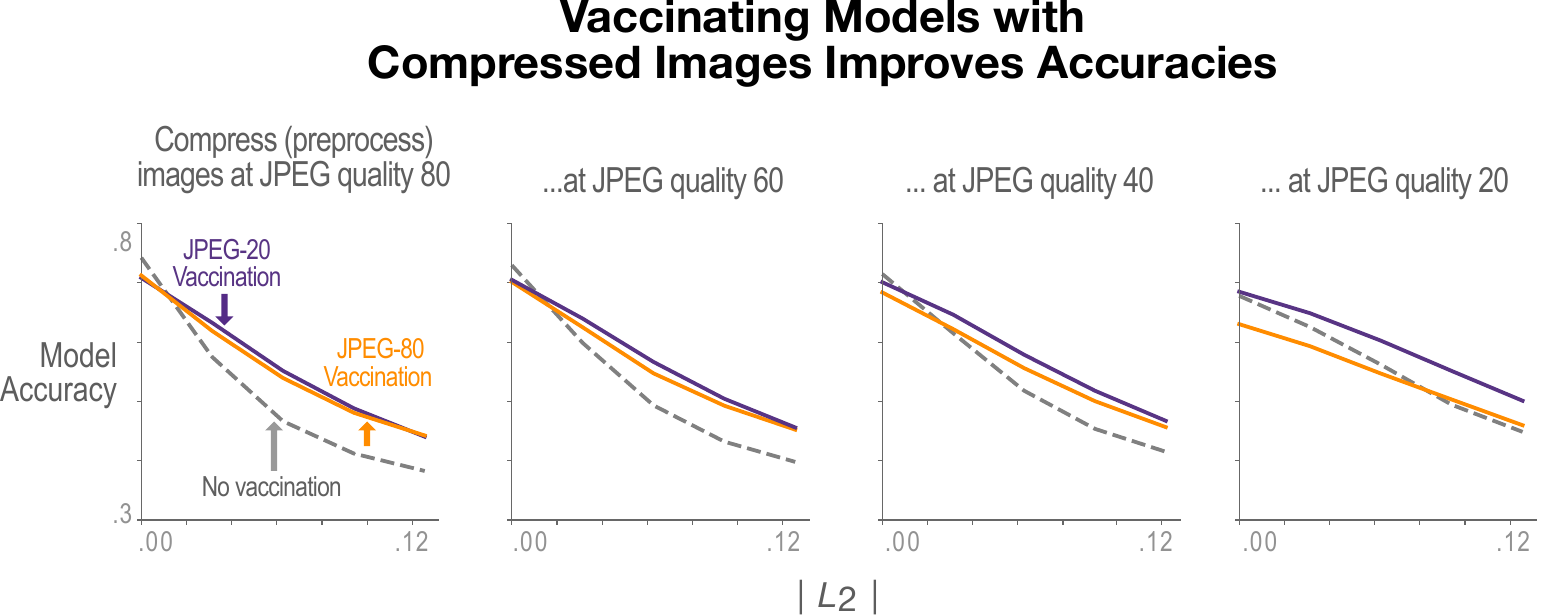}
    \caption{Vaccinating a model by retraining it with compressed images helps recover its accuracy. Each plot shows the model accuracies when preprocessing with different JPEG qualities. Each curve in the plot corresponds to a different model. The gray dotted curve corresponds to the original unvaccinated ResNet-v2 50 model. The orange and purple curves correspond to the models retrained on JPEG qualities 80 and 20 respectively. Retraining on JPEG compressed images and applying JPEG preprocessing helps recover accuracy in a gray-box attack.}
    \label{fig:retrain}
\end{figure*}

We denote the original ResNet-v2 50 model as $\mathcal{M}$, which the adversary has access to.
By retraining on images of a particular JPEG compression quality $q$, we transform $\mathcal{M}$ to $\mathcal{M}_q$, e.g., for JPEG-20 Vaccination, we retrain $\mathcal{M}$ on JPEG-compressed images at quality 20 and obtain $\mathcal{M}_{20}$.
When retraining the ResNet-v2 50 models, we used stochastic gradient descent (SGD) with a learning rate of $5 \times 10^{-3}$, with a decay of 94\% over $~25 \times 10^4$ iterations.
We conducted the retraining on a GPU cluster with 12 NVIDIA Tesla K80 GPUs.
In this manner, we obtain 8 models from quality 20 through quality 90 in increments of 10 ($\mathcal{M}_{20}, \mathcal{M}_{30}, \mathcal{M}_{40} ... \mathcal{M}_{90}$), to cover a wide spectrum of JPEG qualities.
Figure~\ref{fig:retrain} shows the results of model vaccination against FGSM attacks, whose parameter $\epsilon$ ranges from 0 (no perturbation) to 8 (severe perturbation), in steps of 2.
%
The plots show that retraining the model helps   recover even more model accuracy than using JPEG preprocessing alone
(compare the \textit{unvaccinated} gray dotted curve  vs. the \textit{vaccinated} orange and purple curves in Figure~\ref{fig:retrain}).
We found that a given model $\mathcal{M}_q$ performed best when tested with JPEG-compressed images of the same quality $q$, which was expected.

We test these models in an ensemble with two different voting schemes.
The first ensemble scheme, denoted as $\mathcal{M}_q \times q$, corresponds to each model $\mathcal{M}_q$ casting a vote on every JPEG quality $q$ from $q \in \{20, 30, 40, ..., 90 \}$. 
This has a total cost of 64 votes, from which we derive the majority vote. 
In the second scheme, denoted by $\mathcal{M}_q - q$, each model $\mathcal{M}_q$ votes only on $q$, the JPEG quality it was trained on. This incurs a cost of 8 votes.

Table \ref{tab:ensemble} compares the accuracies (against FGSM) and computation costs of these two schemes with those of \ourmethod, which also utilizes an ensemble ($\mathcal{M}_{20}, \mathcal{M}_{40}, \mathcal{M}_{60}, \mathcal{M}_{80}$) with a total of 4 votes. 
\ourmethod achieves very similar performance as compared to the vaccinated models, at half the cost when compared to $\mathcal{M}_q - q$. Hence, \ourmethod offers a favorable trade-off in terms of scalability with minimal effect on accuracy.


\begin{table}[b!]
    \centering
    \begin{tabular}{lrrrrrr}
        \toprule
          Ensemble & Cost & $\epsilon = 0$ & $\epsilon = 2$ & $\epsilon = 4$ & $\epsilon = 6$ & $\epsilon = 8$ \\
        \midrule
          $\mathcal{M}_q \times q$ & 64 & 73.90 & 67.72 & 60.13 & 54.44 & 49.84 \\
        $\mathcal{M}_q - q$      & 8  & 73.54 & 67.06 & 59.86 & 53.91 & 49.40 \\
            \ourmethod               & \textbf{4}  & 72.11 & 66.30 & 59.29 & 53.60 & 48.63  \\
        \bottomrule
    \end{tabular}
    \caption{Comparison of two ensemble schemes with \ourmethod, when defending against FGSM with ResNet-v2 50. 
    $\mathcal{M}_q \times q$ corresponds to each model $\mathcal{M}_q$ voting on each JPEG quality $q$ from $q \in \{20, 30, 40, ..., 90 \}$. In $\mathcal{M}_q - q$, each model $\mathcal{M}_q$ votes only on $q$, the JPEG quality it was trained on. \ourmethod offers a favorable trade-off, providing at least 2x speed-up as compared to larger ensembles, while delivering comparable accuracies.
    }
    \label{tab:ensemble}
\end{table}


\subsection{Transferability in Black-Box Setting}\label{sec:blackbox}
In this setup, we evaluate the transferability of attacked images generated using ResNet-v2 50 on ResNet-v2 101 and Inception-v4.
The attacked images are preprocessed using JPEG compression and Stochastic Local Quantization. 
In Table \ref{tab:blackbox}, we show that JPEG compression as a defense does not significantly reduce model accuracies on low perturbation attacks like DF and CW-L2. 
For higher-perturbation attacks, the accuracy of Inception-v4 lowers by a maximum of 10\%.

\begin{table}[tb]
    \centering
    \begin{tabular}{llrrrr}
        \toprule
         & & \multicolumn{2}{r}{\textbf{Inc-v4} (80.2\%)} & \multicolumn{2}{r}{\textbf{RN-v2 101} (77.0\%)} \\
         Attack & Defense & Accuracy & (Qual.) & Accuracy & (Qual.) \\
        \midrule
        None & JPEG       & 79.05 & (100) & 76.48 & (100)\\
         & SLQ &         75.90 & - &  73.70 & - \gap \\
        CW-L2 & JPEG       & 79.00 & (100) & 76.20 & (100)\\
         & SLQ &     75.80     & - &  73.60 & - \gap\\
        DF & JPEG       & 78.91 & (100) & 76.19 & (100)\\
         & SLQ &     76.29     & - &  73.70 & - \gap \\
        I-FGSM  & JPEG       & 74.84 & (100) & 70.06 & (70)\\
        & SLQ &     73.20     & - & 69.40 & -\gap \\
        FGSM  & JPEG       & 71.00 & (100) & 64.18 & (40)\\
         & SLQ &     70.01     & - & 64.64 & -\\
        \bottomrule
    \end{tabular}
    \caption{
    JPEG compression as defense does not reduce model accuracy significantly on transferred attacks with low perturbation. 
    Adversarial images crafted using the ResNet-v2 50 model are defended using \textit{JPEG} and \textit{Stochastic Local Quantization} (SLQ), before being fed into two other models: Inception-v4 (Inc-v4) and ResNet-v2 101 (RN-v2 101).
    %
    } 
    \label{tab:blackbox}
\end{table}






%
%




\subsection{NIPS 2017 Competition Results}
In addition to the experiment results shown above, 
we also participated in the NIPS 2017 competition on Defense Against Adversarial Attack using a version of our approach that did not include \textit{stochastic local quantization} and \textit{vaccination} to defend against attacks ``in the wild.''
With only an ensemble of three JPEG compression qualities (90, 80, 70),
our entry received a silver badge in the competition, ranking 16th out of more than 100 submissions.

\section{Significance and Impact}
\label{sec:impact}
This work has been making multiple positive impacts on
Intel's research and product development plans.
In this section, we describe such impacts in detail, 
and also describe how they may more broadly influence deep learning and cybersecurity.
We then discuss our work's scope, limitations, and additional practical considerations.

\subsection{Software and Hardware Integration Milestones}


As seen in Section~\ref{sec:eval}, JPEG compression is much faster than other popular preprocessing techniques;
even commodity implementations from Pillow are fast.
However, in order to be deployed into a real defense pipeline, we need to evaluate its computational efficiency with tighter software and hardware integration.
Fortunately, JPEG compression is a widely-used and mature technique that can be be easily deployed in various platforms, and due to its widespread usage, we can use off-the-shelf optimized software and hardware for such testing.
One promising milestone we reached, utilized Intel's hardware Quick Sync Video (QSV) technology: a hardware core dedicated and optimized for video encoding and decoding. 
It was introduced with Sandy Bridge CPU microarchitecture and exists currently in various Intel platforms. 
From our experiments, JPEG compression by Intel QSV is up to 24 times faster
than the Pillow and TensorFlow implementations when evaluated on the same ImageNet validation set of 50,000 images.
This computational efficiency is desirable for applications that need real-time defense, such as autonomous vehicles.
In the future, we plan to explore the feasibility of our approach on more hardware platforms, such as the Intel Movidius Compute Stick\footnote{\url{https://developer.movidius.com}}, which is a low power USB-based deep learning inference kit.

\subsection{New Computational Paradigm: Secure Deep Learning}
This research has sparked insightful discussion with teams of Intel QSV, Intel Deep Learning SDK, and Intel Movidius Compute Stick.
This work not only educates industry regarding concepts and defenses of adversarial machine learning, but also provides opportunities to advance deep learning software and hardware development to incorporate adversarial machine learning defenses.
For example, almost all defenses incur certain levels of computational overhead. 
This may be due to image preprocessing techniques~\cite{guo2018countering, luo2015foveation}, using multiple models for model ensembles~\cite{strauss2017ensemble},
the introduction of adversarial perturbation detectors~\cite{metzen2017detecting, xu2018feature}, or the increase in training time for adversarial training~\cite{goodfellow2014explaining}.
However, while hardware and system improvement for fast deep learning training and inference remains an active area of research,
secure machine learning workloads still receive relatively less attention, suggesting room for improvement.
We believe this will accelerate the positive shift of thinking in the industry in the near future, from addressing problems like 
\textit{``How do we build deep learning accelerators?''} to problems such as 
\textit{``How do we build deep learning accelerators that are not only fast but also secure?''}. 
Understanding such hardware implications are important for microprocessor manufacturers, equipment vendors and companies offering cloud computing services.

\subsection{Scope and Limitations}
In this work, we focus on systematically studying the benefit of compression on its own.
As myriads of newer and stronger attack strategies are continuously discovered, limitations in existing, single defenses are revealed.
Our approach is not a panacea to defend all possible (future) attacks, and we do not expect or intend for it to be used in isolation of other techniques. 
Rather, our methods should be used together with other defense techniques, to potentially develop even stronger defense.
Using multi-layered protection is a proven, long-standing defense strategy that has been pervasive in security research and in practice \cite{tamersoy2014guilt, chen2017predicting}.
Fortunately, since our methods are a preprocessing technique, it is easy to integrate them into many defense pipelines.

\section{Related Work}
\label{sec:related}

Due to intriguing theoretical properties and practical importance, there has been a surge in the number of papers in the past few years attempting to 
find countermeasures against adversarial attacks. 
These include detecting adversarial examples before performing classification~\cite{metzen2017detecting,feinman2017detecting}, modifying network architecture and the underlying primitives used~\cite{gu2014towards,krotov2017dense,ranjan2017improving}, modifying the training process~\cite{goodfellow2014explaining,papernot2016distillation}, and using preprocessing techniques to remove adversarial perturbations~\cite{dziugaite2016study,bhagoji2017dimensionality,luo2015foveation,guo2018countering}.
The preprocessing approach is most relevant to our work.
Below, we describe two methods in this category---median filter and total variation denoising, which we compared against in Section~\ref{sec:eval}.
We then discuss some recent attacks that claim to break preprocessing defenses.



\subsection{Image Preprocessing as Defense}


\textbf{Median Filter}. This method uses a sliding window over the image and replaces each pixel with the median value of its neighboring pixels to spatially smooth the image.
The size of the the sliding window controls the smoothness, for example, a larger window size produces blurrier images.
This technique has been used in multiple prior defense works~\cite{guo2018countering, xu2018feature}.


\medskip
\noindent \textbf{Total Variation Denoising}. The method is based on the principle that images with higher levels of (adversarial) noise tend to have larger total variations: the sum of the absolute difference between adjacent pixel values.
Denoising is performed by reducing the total variation while keeping the denoised image close to the original one.
A weighting parameter is used as a trade-off between the level of total variation and the distance from the original image.
Compared with median filter, this method is more effective at removing adversarial noise while preserving image details~\cite{guo2018countering}.


\subsection{Attacks against Preprocessing Techniques}


One of the reasons why adding preprocessing steps increases attack difficulty is that many preprocessing operations are non-differentiable, thus restricting the feasibility of gradient-based attacks.
In JPEG compression, the quantization in the frequency domain is a non-differentiable operation.

Shin and Song~\cite{shin2017jpeg} propose a method that approximates the quantization in JPEG with a differentiable function.
They also optimize the perturbation over multiple compression qualities to ensure an adversarial image is robust at test time. 
However, the paper only reports preliminary results on 1000 images.
It is also unclear whether their attack is effective against our more advanced \ourmethod method, which introduces more randomization to combat against adversarial noise.

Backward Pass Differentiable Approximation~\cite{athalye2018bbfuscated} is another potential approach to bypass non-differentiable preprocessing techniques.
To attack JPEG preprocessing, it performs forward propagation through the JPEG compression and DNN combination but ignores the compression operation during the backward pass.
This is based on the intuition that the compressed image should look similar to the original one, so the operation can be approximated by the identity function.

\section{Conclusion}
\label{sec:conclusion}

In this paper, we highlighted the urgent need for practical defense for deep learning models that can be readily deployed.
We drew inspiration from JPEG image compression, a well-known and ubiquitous image processing technique, and placed it at the core of our new deep learning model defense framework: \ourmethod.
Since many attack strategies aim to perturb image pixels in ways that are visually imperceptible, the \ourmethod defense framework utilizes JPEG compression to effectively ``compress away'' such pixel manipulation.
\ourmethod immunizes DNN models from being confused by compression artifacts by ``vaccinating'' a model: re-training it with compressed images, where different compression levels are applied to generate multiple vaccinated models that are ultimately used together in an ensemble defense.
Furthermore, \ourmethod adds an additional layer of protection by employing randomization at test time by compressing different regions of an image using random compression levels, making it harder for an adversary to estimate the transformation performed.
This novel combination of vaccination, ensembling and randomization makes \ourmethod a fortified multi-pronged protection, while remaining fast and successful without requiring knowledge about the model.
We conducted extensive, large-scale experiments using the ImageNet dataset, and showed that our approaches eliminate up to 94\% of black-box attacks and 98\% of gray-box attacks delivered by the recent, strongest attacks.
To ensure reproducibility of our results, we have open-sourced our code on GitHub.

\section*{Acknowledgement}
This material is based in part upon work supported by the 
National Science Foundation under Grant Numbers
IIS-1563816, 
CNS-1704701,
and TWC-1526254.
Any opinions, findings, and conclusions or recommendations expressed in this material are those of the author(s) and do not necessarily
 reflect the views of the National Science Foundation.
This research is also supported in part by gifts from Intel, Google, Symantec, Yahoo! Labs, eBay, Amazon, and LogicBlox.

\bibliographystyle{ACM-Reference-Format}
\bibliography{bibliography}


\begin{thebibliography}{35}


\ifx \showCODEN    \undefined \def \showCODEN     #1{\unskip}     \fi
\ifx \showDOI      \undefined \def \showDOI       #1{#1}\fi
\ifx \showISBNx    \undefined \def \showISBNx     #1{\unskip}     \fi
\ifx \showISBNxiii \undefined \def \showISBNxiii  #1{\unskip}     \fi
\ifx \showISSN     \undefined \def \showISSN      #1{\unskip}     \fi
\ifx \showLCCN     \undefined \def \showLCCN      #1{\unskip}     \fi
\ifx \shownote     \undefined \def \shownote      #1{#1}          \fi
\ifx \showarticletitle \undefined \def \showarticletitle #1{#1}   \fi
\ifx \showURL      \undefined \def \showURL       {\relax}        \fi
\providecommand\bibfield[2]{#2}
\providecommand\bibinfo[2]{#2}
\providecommand\natexlab[1]{#1}
\providecommand\showeprint[2][]{arXiv:#2}

\bibitem[\protect\citeauthoryear{Athalye, Carlini, and Wagner}{Athalye
  et~al\mbox{.}}{2018}]%
        {athalye2018bbfuscated}
\bibfield{author}{\bibinfo{person}{Anish Athalye}, \bibinfo{person}{Nicholas
  Carlini}, {and} \bibinfo{person}{David Wagner}.}
  \bibinfo{year}{2018}\natexlab{}.
\newblock \showarticletitle{Obfuscated Gradients Give a False Sense of
  Security: Circumventing Defenses to Adversarial Examples}.
\newblock \bibinfo{journal}{\emph{arXiv preprint arXiv:1802.00420}}
  (\bibinfo{year}{2018}).
\newblock


\bibitem[\protect\citeauthoryear{Athalye and Sutskever}{Athalye and
  Sutskever}{2017}]%
        {athalye2017synthesizing}
\bibfield{author}{\bibinfo{person}{Anish Athalye} {and} \bibinfo{person}{Ilya
  Sutskever}.} \bibinfo{year}{2017}\natexlab{}.
\newblock \showarticletitle{Synthesizing robust adversarial examples}.
\newblock \bibinfo{journal}{\emph{arXiv preprint arXiv:1707.07397}}
  (\bibinfo{year}{2017}).
\newblock


\bibitem[\protect\citeauthoryear{Bhagoji, Cullina, and Mittal}{Bhagoji
  et~al\mbox{.}}{2017}]%
        {bhagoji2017dimensionality}
\bibfield{author}{\bibinfo{person}{Arjun~Nitin Bhagoji},
  \bibinfo{person}{Daniel Cullina}, {and} \bibinfo{person}{Prateek Mittal}.}
  \bibinfo{year}{2017}\natexlab{}.
\newblock \showarticletitle{Dimensionality Reduction as a Defense against
  Evasion Attacks on Machine Learning Classifiers}.
\newblock \bibinfo{journal}{\emph{arXiv preprint arXiv:1704.02654}}
  (\bibinfo{year}{2017}).
\newblock


\bibitem[\protect\citeauthoryear{Carlini and Wagner}{Carlini and
  Wagner}{2017}]%
        {carlini2017towards}
\bibfield{author}{\bibinfo{person}{Nicholas Carlini} {and}
  \bibinfo{person}{David Wagner}.} \bibinfo{year}{2017}\natexlab{}.
\newblock \showarticletitle{Towards evaluating the robustness of neural
  networks}. In \bibinfo{booktitle}{\emph{Security and Privacy (SP), 2017 IEEE
  Symposium on}}. IEEE, \bibinfo{pages}{39--57}.
\newblock


\bibitem[\protect\citeauthoryear{Chen, Han, Chau, Gates, Hart, and Roundy}{Chen
  et~al\mbox{.}}{2017}]%
        {chen2017predicting}
\bibfield{author}{\bibinfo{person}{Shang-Tse Chen}, \bibinfo{person}{Yufei
  Han}, \bibinfo{person}{Duen~Horng Chau}, \bibinfo{person}{Christopher Gates},
  \bibinfo{person}{Michael Hart}, {and} \bibinfo{person}{Kevin~A Roundy}.}
  \bibinfo{year}{2017}\natexlab{}.
\newblock \showarticletitle{Predicting Cyber Threats with Virtual Security
  Products}. In \bibinfo{booktitle}{\emph{Proceedings of the 33rd Annual
  Computer Security Applications Conference}}. ACM, \bibinfo{pages}{189--199}.
\newblock


\bibitem[\protect\citeauthoryear{Chuan~Guo}{Chuan~Guo}{2018}]%
        {guo2018countering}
\bibfield{author}{\bibinfo{person}{Moustapha Cisse Laurens van der~Maaten
  Chuan~Guo, Mayank~Rana}.} \bibinfo{year}{2018}\natexlab{}.
\newblock \showarticletitle{Countering Adversarial Images using Input
  Transformations}.
\newblock \bibinfo{journal}{\emph{International Conference on Learning
  Representations}} (\bibinfo{year}{2018}).
\newblock
\urldef\tempurl%
\url{https://openreview.net/forum?id=SyJ7ClWCb}
\showURL{%
\tempurl}


\bibitem[\protect\citeauthoryear{Das, Shanbhogue, Chen, Hohman, Chen, Kounavis,
  and Chau}{Das et~al\mbox{.}}{2017}]%
        {das2017keeping}
\bibfield{author}{\bibinfo{person}{Nilaksh Das}, \bibinfo{person}{Madhuri
  Shanbhogue}, \bibinfo{person}{Shang-Tse Chen}, \bibinfo{person}{Fred Hohman},
  \bibinfo{person}{Li Chen}, \bibinfo{person}{Michael~E Kounavis}, {and}
  \bibinfo{person}{Duen~Horng Chau}.} \bibinfo{year}{2017}\natexlab{}.
\newblock \showarticletitle{Keeping the bad guys out: Protecting and
  vaccinating deep learning with jpeg compression}.
\newblock \bibinfo{journal}{\emph{arXiv preprint arXiv:1705.02900}}
  (\bibinfo{year}{2017}).
\newblock


\bibitem[\protect\citeauthoryear{Dziugaite, Ghahramani, and Roy}{Dziugaite
  et~al\mbox{.}}{2016}]%
        {dziugaite2016study}
\bibfield{author}{\bibinfo{person}{Gintare~Karolina Dziugaite},
  \bibinfo{person}{Zoubin Ghahramani}, {and} \bibinfo{person}{Daniel~M Roy}.}
  \bibinfo{year}{2016}\natexlab{}.
\newblock \showarticletitle{A study of the effect of JPG compression on
  adversarial images}.
\newblock \bibinfo{journal}{\emph{arXiv preprint arXiv:1608.00853}}
  (\bibinfo{year}{2016}).
\newblock


\bibitem[\protect\citeauthoryear{Evtimov, Eykholt, Fernandes, Kohno, Li,
  Prakash, Rahmati, and Song}{Evtimov et~al\mbox{.}}{2017}]%
        {evtimov2017robust}
\bibfield{author}{\bibinfo{person}{Ivan Evtimov}, \bibinfo{person}{Kevin
  Eykholt}, \bibinfo{person}{Earlence Fernandes}, \bibinfo{person}{Tadayoshi
  Kohno}, \bibinfo{person}{Bo Li}, \bibinfo{person}{Atul Prakash},
  \bibinfo{person}{Amir Rahmati}, {and} \bibinfo{person}{Dawn Song}.}
  \bibinfo{year}{2017}\natexlab{}.
\newblock \showarticletitle{Robust physical-world attacks on machine learning
  models}.
\newblock \bibinfo{journal}{\emph{arXiv preprint arXiv:1707.08945}}
  (\bibinfo{year}{2017}).
\newblock


\bibitem[\protect\citeauthoryear{Eykholt, Evtimov, Fernandes, Li, Song, Kohno,
  Rahmati, Prakash, and Tramer}{Eykholt et~al\mbox{.}}{2017}]%
        {eykholt2017note}
\bibfield{author}{\bibinfo{person}{Kevin Eykholt}, \bibinfo{person}{Ivan
  Evtimov}, \bibinfo{person}{Earlence Fernandes}, \bibinfo{person}{Bo Li},
  \bibinfo{person}{Dawn Song}, \bibinfo{person}{Tadayoshi Kohno},
  \bibinfo{person}{Amir Rahmati}, \bibinfo{person}{Atul Prakash}, {and}
  \bibinfo{person}{Florian Tramer}.} \bibinfo{year}{2017}\natexlab{}.
\newblock \showarticletitle{Note on Attacking Object Detectors with Adversarial
  Stickers}.
\newblock \bibinfo{journal}{\emph{arXiv preprint arXiv:1712.08062}}
  (\bibinfo{year}{2017}).
\newblock


\bibitem[\protect\citeauthoryear{Feinman, Curtin, Shintre, and Gardner}{Feinman
  et~al\mbox{.}}{2017}]%
        {feinman2017detecting}
\bibfield{author}{\bibinfo{person}{Reuben Feinman}, \bibinfo{person}{Ryan~R
  Curtin}, \bibinfo{person}{Saurabh Shintre}, {and} \bibinfo{person}{Andrew~B
  Gardner}.} \bibinfo{year}{2017}\natexlab{}.
\newblock \showarticletitle{Detecting Adversarial Samples from Artifacts}.
\newblock \bibinfo{journal}{\emph{arXiv preprint arXiv:1703.00410}}
  (\bibinfo{year}{2017}).
\newblock


\bibitem[\protect\citeauthoryear{Goodfellow, Shlens, and Szegedy}{Goodfellow
  et~al\mbox{.}}{2014}]%
        {goodfellow2014explaining}
\bibfield{author}{\bibinfo{person}{Ian~J Goodfellow}, \bibinfo{person}{Jonathon
  Shlens}, {and} \bibinfo{person}{Christian Szegedy}.}
  \bibinfo{year}{2014}\natexlab{}.
\newblock \showarticletitle{Explaining and harnessing adversarial examples}. In
  \bibinfo{booktitle}{\emph{ICLR}}.
\newblock


\bibitem[\protect\citeauthoryear{Grosse, Papernot, Manoharan, Backes, and
  McDaniel}{Grosse et~al\mbox{.}}{2016}]%
        {grosse2016malware}
\bibfield{author}{\bibinfo{person}{Kathrin Grosse}, \bibinfo{person}{Nicolas
  Papernot}, \bibinfo{person}{Praveen Manoharan}, \bibinfo{person}{Michael
  Backes}, {and} \bibinfo{person}{Patrick McDaniel}.}
  \bibinfo{year}{2016}\natexlab{}.
\newblock \showarticletitle{Adversarial perturbations against deep neural
  networks for malware classification}.
\newblock \bibinfo{journal}{\emph{arXiv preprint arXiv:1606.04435}}
  (\bibinfo{year}{2016}).
\newblock


\bibitem[\protect\citeauthoryear{Gu and Rigazio}{Gu and Rigazio}{2014}]%
        {gu2014towards}
\bibfield{author}{\bibinfo{person}{Shixiang Gu} {and} \bibinfo{person}{Luca
  Rigazio}.} \bibinfo{year}{2014}\natexlab{}.
\newblock \showarticletitle{Towards deep neural network architectures robust to
  adversarial examples}.
\newblock \bibinfo{journal}{\emph{arXiv preprint arXiv:1412.5068}}
  (\bibinfo{year}{2014}).
\newblock


\bibitem[\protect\citeauthoryear{Hu and Tan}{Hu and Tan}{2017}]%
        {hu2017generating}
\bibfield{author}{\bibinfo{person}{Weiwei Hu} {and} \bibinfo{person}{Ying
  Tan}.} \bibinfo{year}{2017}\natexlab{}.
\newblock \showarticletitle{Generating Adversarial Malware Examples for
  Black-Box Attacks Based on GAN}.
\newblock \bibinfo{journal}{\emph{arXiv preprint arXiv:1702.05983}}
  (\bibinfo{year}{2017}).
\newblock


\bibitem[\protect\citeauthoryear{Huang, Papernot, Goodfellow, Duan, and
  Abbeel}{Huang et~al\mbox{.}}{2017}]%
        {huang2017adversarial}
\bibfield{author}{\bibinfo{person}{Sandy Huang}, \bibinfo{person}{Nicolas
  Papernot}, \bibinfo{person}{Ian Goodfellow}, \bibinfo{person}{Yan Duan},
  {and} \bibinfo{person}{Pieter Abbeel}.} \bibinfo{year}{2017}\natexlab{}.
\newblock \showarticletitle{Adversarial attacks on neural network policies}.
\newblock \bibinfo{journal}{\emph{arXiv preprint arXiv:1702.02284}}
  (\bibinfo{year}{2017}).
\newblock


\bibitem[\protect\citeauthoryear{Krizhevsky, Sutskever, and Hinton}{Krizhevsky
  et~al\mbox{.}}{2012}]%
        {krizhevsky2012imagenet}
\bibfield{author}{\bibinfo{person}{Alex Krizhevsky}, \bibinfo{person}{Ilya
  Sutskever}, {and} \bibinfo{person}{Geoffrey~E Hinton}.}
  \bibinfo{year}{2012}\natexlab{}.
\newblock \showarticletitle{Imagenet classification with deep convolutional
  neural networks}. In \bibinfo{booktitle}{\emph{Advances in neural information
  processing systems}}. \bibinfo{pages}{1097--1105}.
\newblock


\bibitem[\protect\citeauthoryear{Krotov and Hopfield}{Krotov and
  Hopfield}{2017}]%
        {krotov2017dense}
\bibfield{author}{\bibinfo{person}{Dmitry Krotov} {and} \bibinfo{person}{John~J
  Hopfield}.} \bibinfo{year}{2017}\natexlab{}.
\newblock \showarticletitle{Dense Associative Memory is Robust to Adversarial
  Inputs}.
\newblock \bibinfo{journal}{\emph{arXiv preprint arXiv:1701.00939}}
  (\bibinfo{year}{2017}).
\newblock


\bibitem[\protect\citeauthoryear{Kurakin, Goodfellow, and Bengio}{Kurakin
  et~al\mbox{.}}{2016}]%
        {kurakin2016adversarial}
\bibfield{author}{\bibinfo{person}{Alexey Kurakin}, \bibinfo{person}{Ian
  Goodfellow}, {and} \bibinfo{person}{Samy Bengio}.}
  \bibinfo{year}{2016}\natexlab{}.
\newblock \showarticletitle{Adversarial examples in the physical world}.
\newblock \bibinfo{journal}{\emph{arXiv preprint arXiv:1607.02533}}
  (\bibinfo{year}{2016}).
\newblock


\bibitem[\protect\citeauthoryear{Lin, Hong, Liao, Shih, Liu, and Sun}{Lin
  et~al\mbox{.}}{2017}]%
        {lin2017tactics}
\bibfield{author}{\bibinfo{person}{Yen-Chen Lin}, \bibinfo{person}{Zhang-Wei
  Hong}, \bibinfo{person}{Yuan-Hong Liao}, \bibinfo{person}{Meng-Li Shih},
  \bibinfo{person}{Ming-Yu Liu}, {and} \bibinfo{person}{Min Sun}.}
  \bibinfo{year}{2017}\natexlab{}.
\newblock \showarticletitle{Tactics of Adversarial Attack on Deep Reinforcement
  Learning Agents}.
\newblock \bibinfo{journal}{\emph{arXiv preprint arXiv:1703.06748}}
  (\bibinfo{year}{2017}).
\newblock


\bibitem[\protect\citeauthoryear{Luo, Boix, Roig, Poggio, and Zhao}{Luo
  et~al\mbox{.}}{2015}]%
        {luo2015foveation}
\bibfield{author}{\bibinfo{person}{Yan Luo}, \bibinfo{person}{Xavier Boix},
  \bibinfo{person}{Gemma Roig}, \bibinfo{person}{Tomaso Poggio}, {and}
  \bibinfo{person}{Qi Zhao}.} \bibinfo{year}{2015}\natexlab{}.
\newblock \showarticletitle{Foveation-based mechanisms alleviate adversarial
  examples}.
\newblock \bibinfo{journal}{\emph{arXiv preprint arXiv:1511.06292}}
  (\bibinfo{year}{2015}).
\newblock


\bibitem[\protect\citeauthoryear{Metzen, Genewein, Fischer, and
  Bischoff}{Metzen et~al\mbox{.}}{2017}]%
        {metzen2017detecting}
\bibfield{author}{\bibinfo{person}{Jan~Hendrik Metzen}, \bibinfo{person}{Tim
  Genewein}, \bibinfo{person}{Volker Fischer}, {and} \bibinfo{person}{Bastian
  Bischoff}.} \bibinfo{year}{2017}\natexlab{}.
\newblock \showarticletitle{On detecting adversarial perturbations}. In
  \bibinfo{booktitle}{\emph{ICLR}}.
\newblock


\bibitem[\protect\citeauthoryear{Moosavi~Dezfooli, Fawzi, Fawzi, and
  Frossard}{Moosavi~Dezfooli et~al\mbox{.}}{2017}]%
        {Moosavi17}
\bibfield{author}{\bibinfo{person}{Seyed~Mohsen Moosavi~Dezfooli},
  \bibinfo{person}{Alhussein Fawzi}, \bibinfo{person}{Omar Fawzi}, {and}
  \bibinfo{person}{Pascal Frossard}.} \bibinfo{year}{2017}\natexlab{}.
\newblock \showarticletitle{Universal adversarial perturbations}. In
  \bibinfo{booktitle}{\emph{CVPR}}.
\newblock


\bibitem[\protect\citeauthoryear{Moosavi-Dezfooli, Fawzi, and
  Frossard}{Moosavi-Dezfooli et~al\mbox{.}}{2016}]%
        {Moosavi16}
\bibfield{author}{\bibinfo{person}{Seyed-Mohsen Moosavi-Dezfooli},
  \bibinfo{person}{Alhussein Fawzi}, {and} \bibinfo{person}{Pascal Frossard}.}
  \bibinfo{year}{2016}\natexlab{}.
\newblock \showarticletitle{DeepFool: A Simple and Accurate Method to Fool Deep
  Neural Networks}. In \bibinfo{booktitle}{\emph{CVPR}}.
\newblock


\bibitem[\protect\citeauthoryear{Papernot, McDaniel, Goodfellow, Jha, Celik,
  and Swami}{Papernot et~al\mbox{.}}{2017}]%
        {Papernot17blackbox}
\bibfield{author}{\bibinfo{person}{Nicolas Papernot}, \bibinfo{person}{Patrick
  McDaniel}, \bibinfo{person}{Ian Goodfellow}, \bibinfo{person}{Somesh Jha},
  \bibinfo{person}{Z.~Berkay Celik}, {and} \bibinfo{person}{Ananthram Swami}.}
  \bibinfo{year}{2017}\natexlab{}.
\newblock \showarticletitle{Practical Black-Box Attacks Against Machine
  Learning}. In \bibinfo{booktitle}{\emph{Proceedings of the 2017 ACM on Asia
  Conference on Computer and Communications Security}}
  \emph{(\bibinfo{series}{ASIA CCS '17})}. \bibinfo{pages}{506--519}.
\newblock


\bibitem[\protect\citeauthoryear{Papernot, McDaniel, Wu, Jha, and
  Swami}{Papernot et~al\mbox{.}}{2016c}]%
        {papernot2016distillation}
\bibfield{author}{\bibinfo{person}{Nicolas Papernot}, \bibinfo{person}{Patrick
  McDaniel}, \bibinfo{person}{Xi Wu}, \bibinfo{person}{Somesh Jha}, {and}
  \bibinfo{person}{Ananthram Swami}.} \bibinfo{year}{2016}\natexlab{c}.
\newblock \showarticletitle{Distillation as a defense to adversarial
  perturbations against deep neural networks}. In
  \bibinfo{booktitle}{\emph{IEEE Symposium on Security and Privacy}}.
  \bibinfo{pages}{582--597}.
\newblock


\bibitem[\protect\citeauthoryear{Papernot, McDaniel, Jha, Fredrikson, Celik,
  and Swami}{Papernot et~al\mbox{.}}{2016a}]%
        {Papernot16limitation}
\bibfield{author}{\bibinfo{person}{Nicolas Papernot},
  \bibinfo{person}{Patrick~D. McDaniel}, \bibinfo{person}{Somesh Jha},
  \bibinfo{person}{Matt Fredrikson}, \bibinfo{person}{Z.~Berkay Celik}, {and}
  \bibinfo{person}{Ananthram Swami}.} \bibinfo{year}{2016}\natexlab{a}.
\newblock \showarticletitle{The Limitations of Deep Learning in Adversarial
  Settings}. In \bibinfo{booktitle}{\emph{{IEEE} European Symposium on Security
  and Privacy, EuroS{\&}P 2016, Saarbr{\"{u}}cken, Germany, March 21-24,
  2016}}. \bibinfo{pages}{372--387}.
\newblock


\bibitem[\protect\citeauthoryear{Papernot, McDaniel, Swami, and
  Harang}{Papernot et~al\mbox{.}}{2016b}]%
        {PapernotMSH16}
\bibfield{author}{\bibinfo{person}{Nicolas Papernot},
  \bibinfo{person}{Patrick~D. McDaniel}, \bibinfo{person}{Ananthram Swami},
  {and} \bibinfo{person}{Richard~E. Harang}.} \bibinfo{year}{2016}\natexlab{b}.
\newblock \showarticletitle{Crafting adversarial input sequences for recurrent
  neural networks}. In \bibinfo{booktitle}{\emph{2016 {IEEE} Military
  Communications Conference, {MILCOM}}}. \bibinfo{pages}{49--54}.
\newblock


\bibitem[\protect\citeauthoryear{Ranjan, Sankaranarayanan, Castillo, and
  Chellappa}{Ranjan et~al\mbox{.}}{2017}]%
        {ranjan2017improving}
\bibfield{author}{\bibinfo{person}{Rajeev Ranjan}, \bibinfo{person}{Swami
  Sankaranarayanan}, \bibinfo{person}{Carlos~D Castillo}, {and}
  \bibinfo{person}{Rama Chellappa}.} \bibinfo{year}{2017}\natexlab{}.
\newblock \showarticletitle{Improving Network Robustness against Adversarial
  Attacks with Compact Convolution}.
\newblock \bibinfo{journal}{\emph{arXiv preprint arXiv:1712.00699}}
  (\bibinfo{year}{2017}).
\newblock


\bibitem[\protect\citeauthoryear{Sharif, Bhagavatula, Bauer, and Reiter}{Sharif
  et~al\mbox{.}}{2016}]%
        {sharif2016accessorize}
\bibfield{author}{\bibinfo{person}{Mahmood Sharif}, \bibinfo{person}{Sruti
  Bhagavatula}, \bibinfo{person}{Lujo Bauer}, {and} \bibinfo{person}{Michael~K
  Reiter}.} \bibinfo{year}{2016}\natexlab{}.
\newblock \showarticletitle{Accessorize to a crime: Real and stealthy attacks
  on state-of-the-art face recognition}. In
  \bibinfo{booktitle}{\emph{Proceedings of the 2016 ACM SIGSAC Conference on
  Computer and Communications Security}}. ACM, \bibinfo{pages}{1528--1540}.
\newblock


\bibitem[\protect\citeauthoryear{Shin and Song}{Shin and Song}{2017}]%
        {shin2017jpeg}
\bibfield{author}{\bibinfo{person}{Richard Shin} {and} \bibinfo{person}{Dawn
  Song}.} \bibinfo{year}{2017}\natexlab{}.
\newblock \showarticletitle{JPEG-resistant Adversarial Images}.
\newblock \bibinfo{journal}{\emph{NIPS 2017 Workshop on Machine Learning and
  Computer Security}} (\bibinfo{year}{2017}).
\newblock


\bibitem[\protect\citeauthoryear{Strauss, Hanselmann, Junginger, and
  Ulmer}{Strauss et~al\mbox{.}}{2017}]%
        {strauss2017ensemble}
\bibfield{author}{\bibinfo{person}{Thilo Strauss}, \bibinfo{person}{Markus
  Hanselmann}, \bibinfo{person}{Andrej Junginger}, {and}
  \bibinfo{person}{Holger Ulmer}.} \bibinfo{year}{2017}\natexlab{}.
\newblock \showarticletitle{Ensemble methods as a defense to adversarial
  perturbations against deep neural networks}.
\newblock \bibinfo{journal}{\emph{arXiv preprint arXiv:1709.03423}}
  (\bibinfo{year}{2017}).
\newblock


\bibitem[\protect\citeauthoryear{Szegedy, Inc, Zaremba, Sutskever, Inc, Bruna,
  Erhan, Inc, Goodfellow, and Fergus}{Szegedy et~al\mbox{.}}{2014}]%
        {Szegedy14}
\bibfield{author}{\bibinfo{person}{Christian Szegedy}, \bibinfo{person}{Google
  Inc}, \bibinfo{person}{Wojciech Zaremba}, \bibinfo{person}{Ilya Sutskever},
  \bibinfo{person}{Google Inc}, \bibinfo{person}{Joan Bruna},
  \bibinfo{person}{Dumitru Erhan}, \bibinfo{person}{Google Inc},
  \bibinfo{person}{Ian Goodfellow}, {and} \bibinfo{person}{Rob Fergus}.}
  \bibinfo{year}{2014}\natexlab{}.
\newblock \showarticletitle{Intriguing properties of neural networks}. In
  \bibinfo{booktitle}{\emph{ICLR}}.
\newblock


\bibitem[\protect\citeauthoryear{Tamersoy, Roundy, and Chau}{Tamersoy
  et~al\mbox{.}}{2014}]%
        {tamersoy2014guilt}
\bibfield{author}{\bibinfo{person}{Acar Tamersoy}, \bibinfo{person}{Kevin
  Roundy}, {and} \bibinfo{person}{Duen~Horng Chau}.}
  \bibinfo{year}{2014}\natexlab{}.
\newblock \showarticletitle{Guilt by association: large scale malware detection
  by mining file-relation graphs}. In \bibinfo{booktitle}{\emph{Proceedings of
  the 20th ACM SIGKDD international conference on Knowledge discovery and data
  mining}}. ACM, \bibinfo{pages}{1524--1533}.
\newblock


\bibitem[\protect\citeauthoryear{Xu, Evans, and Qi}{Xu et~al\mbox{.}}{2018}]%
        {xu2018feature}
\bibfield{author}{\bibinfo{person}{Weilin Xu}, \bibinfo{person}{David Evans},
  {and} \bibinfo{person}{Yanjun Qi}.} \bibinfo{year}{2018}\natexlab{}.
\newblock \showarticletitle{{Feature Squeezing: Detecting Adversarial Examples
  in Deep Neural Networks}}. In \bibinfo{booktitle}{\emph{Proceedings of the
  2018 Network and Distributed Systems Security Symposium (NDSS)}}.
\newblock


\end{thebibliography}

\end{document}